\documentclass[letterpaper, 10 pt, conference]{ieeeconf}

\IEEEoverridecommandlockouts
\usepackage{cite}
\usepackage{float}
\usepackage[pagewise]{lineno}
\usepackage[super]{nth}

\newcommand{\fref}[1]{Fig.~\ref{#1}}
\newcommand{\sref}[1]{Section~\ref{#1}}
\newcommand{\tref}[1]{Table~\ref{#1}}
\newcommand{\fix}[1]{#1}

\usepackage{graphicx}
\graphicspath{{./figures/}}
\usepackage{amsmath}
\usepackage{amsfonts}
\usepackage[amssymb]{SIunits}
\usepackage{booktabs}
\usepackage{multirow}
\usepackage{array}
\usepackage{subfig}
\usepackage{graphicx}
\usepackage{hyperref}
\usepackage[misc,geometry]{ifsym}
\title{\LARGE \bf
AirLoc: Object-based Indoor Relocalization
}

\author{Aryan$^{1}$, Bowen Li$^{2}$, Sebastian Scherer$^{2}$, Yun-Jou Lin$^{3}$, and Chen Wang$^{4}$% <-this % stops a space
\thanks{$^{1}$The Department of Electronics and Communication, Delhi Technological University, Delhi, India
{\tt\small aryanmangal2022@gmail.com}}%
\thanks{$^{2}$The Robotics Institute, Carnegie Mellon University, Pittsburgh, PA 15213, USA
{\tt\small \{bowenli2, basti\}@andrew.cmu.edu}}%
\thanks{$^{3}$OPPO Palo Alto, California, USA. {\tt\small rose.lin@oppo.com}}%
\thanks{$^{4}$The Department of Computer Science and Engineering, State University of New York at Buffalo, NY 14260, USA. {\tt\small chenwang@dr.com}}%
}

\begin{document}
% \linenumbers % Uncomment this to enable line numbers in the peer review

\maketitle
\thispagestyle{empty}
\pagestyle{empty}

%%%%%%%%%%%%%%%%%%%%%%%%%%%%%%%%%%%%%%%%%%%%%%%%%%%%%%%%%%%%%%%%%%%%%%%%%%%%%%%%
\begin{abstract}

Indoor relocalization is vital for both robotic tasks like autonomous exploration and civil applications such as navigation with a cell phone in a shopping mall.
Some previous approaches adopt geometrical information such as key-point features or local textures to carry out indoor relocalization, but they either easily fail in environment with visually similar scenes or require many database images. 
Inspired by the fact that humans often remember places by recognizing unique landmarks, we resort to objects, which are more informative than geometry elements.
In this work, we propose a simple yet effective object-based indoor relocalization approach, dubbed AirLoc.
To overcome the critical challenges of the object reidentification and remembering object relationships, we extract object-wise appearance embedding and inter-object geometric relationship. The geometry and appearance features are integrated to generate cumulative scene features. % 
This results in a robust, accurate, and portable indoor relocalization system, which outperforms the state-of-the-art methods in room-level relocalization by 9.5\% of PR-AUC and 7\% of accuracy.
In addition to exhaustive evaluation, we also carry out real-world tests, where AirLoc shows robustness in challenges like severe occlusion, perceptual aliasing, viewpoint shift, and deformation.

\end{abstract}


%%%%%%%%%%%%%%%%%%%%%%%%%%%%%%%%%%%%%%%%%%%%%%%%%%%%%%%%%%%%%%%%%%%%%%%%%%%%%%%%
\section{INTRODUCTION}

Indoor relocalization has gained increasing attention with the development of numerous mobile phones and robotic applications such as virtual reality (VR) \cite{meng2018exploiting}, augmented reality (AR) \cite{khan2019impact}, and robot navigation \cite{shahjalal2018implementation}.
For example, it can be employed in large buildings such as shopping malls and offices where one can use cell phone for self-relocalization when lost. 
Additionally, many existing mobile robot localization techniques, such as visual odometry \cite{bavle2018stereo} and simultaneous localization and mapping (SLAM) \cite{li2022loop} require indoor relocalization to correct accumulated drift. 

Many algorithms \cite{tian20203d} focus on providing accurate pose estimation, however, exact camera poses are often not required by civil applications such as indoor navigation. For instance, a lost patient in a hospital just wants to figure out which room he is in, instead of the precise centimeter-level location. Besides, we expect to re-recognize a place with only a few image samples (database), making the system commercially viable. If the method requires a large database, generalizing the system and creating such a database for multiple places is impractical due to memory and latency issues.

In recent years, indoor relocalization methods have been focusing on geometric textures with key-point features \cite{mur2015orb,bay2006surf} or semantic information \cite{guo2021semantic}.  
However, they are often not scalable for two major reasons. 
First, they require either a 3D scene model \cite{he2020pvn3d} or a large number of database images \cite{arandjelovic2016netvlad}, which are not readily accessible in most real-world indoor scenes. 
Secondly, these methods can't work well in challenging scenarios like occlusion, light changes, and the interference of dynamic objects such as humans. It is because that they heavily rely on local texture matching which often produces false matches for illumination change or visually similar scenes. 
Image-based methods, such as NetVLAD \cite{arandjelovic2016netvlad} and PatchNetVLAD \cite{hausler2021patch}, also produce false matches because they rely on the collective features of an image rather than understanding the individual identities depicted in the image.

It remains questionable whether these challenges can be resolved with a limited number of database images available. 
Therefore, in this paper we resort to use higher-level information such as objects' appearance and relative geometry to tackle the problem of indoor relocalization.

\begin{figure}[t]
    \centering
    \vspace{7pt}
    \includegraphics[width=\linewidth]{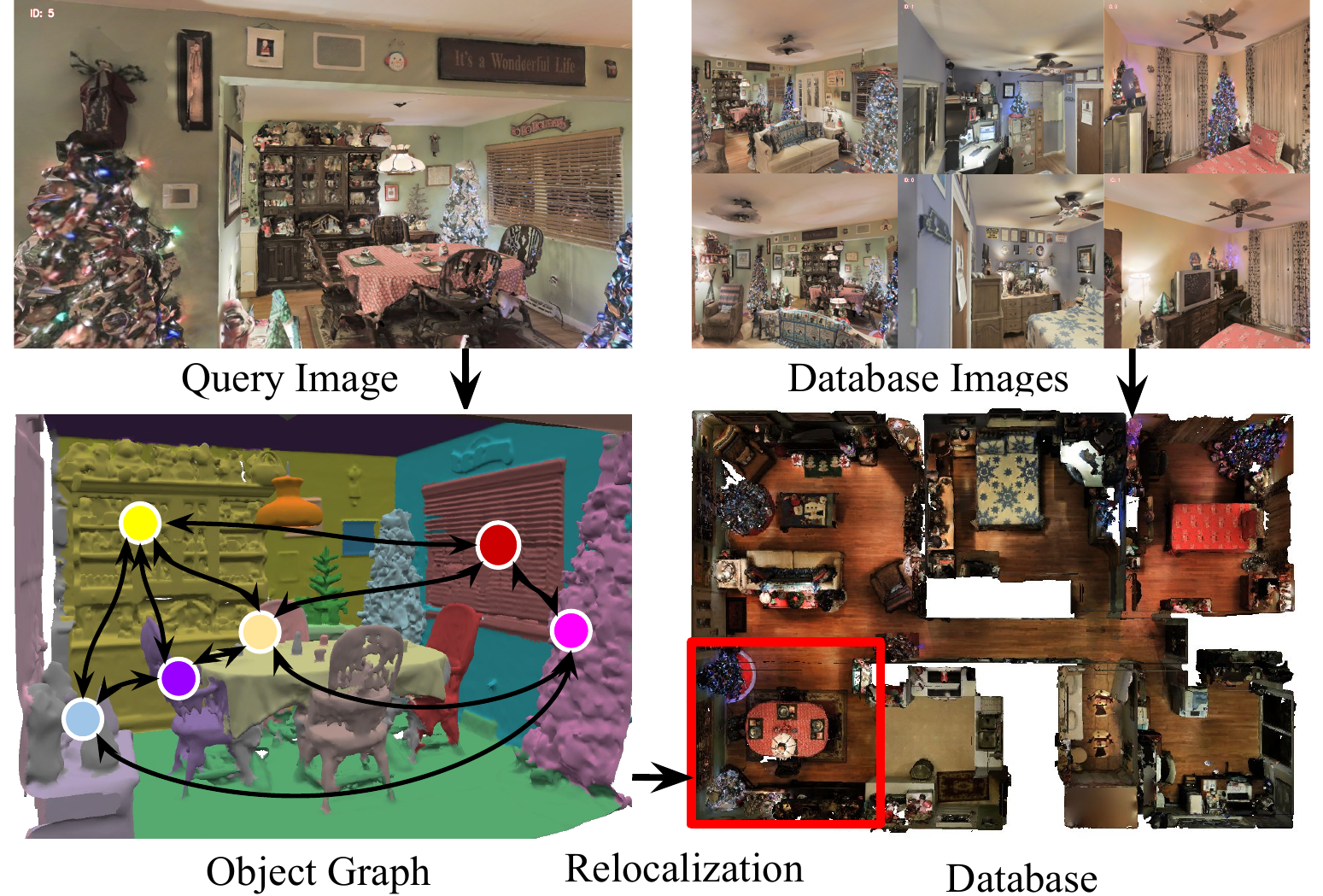}
    \caption{The pipeline of AirLoc for object-based indoor relocalization. AirLoc can provide room-level relocalization by constructing an object graph using one query image comparing with the database, which can be established with only $K$ ($K=1,2,5,10$) images for each room.
    }
    \label{fig:Front_page}
\end{figure}

Researchers have shown increasing interest in object encoding and re-identification tasks \cite{xu2022aircode, keetha2022airobject}. 
The strong representation from objects can be utilized for re-identification with amazing efficacy. 
Inspired by this, we propose AirLoc, an object-based indoor relocalization approach shown in \fref{fig:Front_page}, which fully utilizes appearance and geometry relations.
We present that room-level relocalization for a single query image can be effectively achieved given a database of rooms.
Furthermore, since the model is usually expected to quickly generalize to new environments where a large number of database images can't be quickly obtained, we take only a few images ($K=1,2,3,5,10$) from every room to construct the database. 
AirLoc outperforms various baselines and achieves an amazing speed of 20 \milli\second/frame, making it affordable for low-power mobile robots or cellphones, which demonstrates its outstanding effectiveness and robustness.
In summary, the main contributions of this paper are:

\begin{itemize}
    \item We introduce a simple yet effective indoor relocalization framework, named AirLoc, that relies on object-level information to overcome the limitations of local feature or image-based approaches.
    
    \item We propose two modules to extract appearance and geometry-related features, respectively, which are then combined to perform room-level relocalization.
    \item We perform exhaustive experimental evaluation are on a newly rendered Reloc110 dataset, which contains 306K images and 113 rooms. AirLoc can robustly outperform the state-of-the-art methods, obtaining improvement of 9.5\% PR-AUC and 7\% accuracy.

    \item We implement real-world tests to validate the robustness of AirLoc to illumination change, occlusion, and viewpoint shift. We release source code at 
    \url{https://github.com/sair-lab/AirLoc}
    to benefit the robotics community.
\end{itemize} 

\section{Related Work}

We first review the related datasets for indoor relocalization. Then the methods based on key-point feature \cite{ng2003sift,bay2006surf, arandjelovic2016netvlad } and objects \cite{xu2022aircode, keetha2022airobject} are presented, respectively.

\subsection{Datasets for Indoor Relocalization}
Many datasets have been collected for semantic scene understanding. The Places365-Standard dataset \cite{zhou2017places} is built for visual understanding tasks like scene context, action and event prediction, and object recognition. It contains 1.8 million train images from 365 scene categories.
ADE20k \cite{zhou2017scene} dataset contains images exhaustively annotated with objects and object-parts with additional information of occlusion. 
MIT Indoor scenes database \cite{quattoni2009recognizing}  contains 67 Indoor categories and 15620 images but distribution of images varies per category.  
A recently introduced indoor RGB-D dataset, RIO10 \cite{wald2019rio} has changing indoor environments containing 74 sequences split into training, validation and testing sets.

Datasets used for object-based scene understanding tasks, such as real-world indoor relocalization, should include room labels.
Properties such as ground truth  segmentation and 
images from varying viewpoints
are also important for finer learning. 
Existing datasets miss at least one of the above characteristics, which motivates us to construct a new dataset with such labels and specifications.

\subsection{Key-point and Image feature-based Methods}
Handcrafted key-point features such as SIFT \cite{lowe2004distinctive} and SURF \cite{bay2006surf} have been widely applied to conventional methods such as image retrieval, loop closure detection, and Visual Place Recognition (VPR). 
A binary descriptor ORB \cite{rublee2011orb} was utilized in DBoW2 \cite{galvez2012bags} for image retrieval using visual vocabulary of features.
However, these handcrafted local features are not discriminative in more complex and clutted environments, where the conventional methods easily fail.

Compared to handcrafted features, approaches using deep learned features have been proved more robust \cite{chen2017deep}. SuperPoint \cite{detone2018superpoint}, a recently proposed deep learning method, uses self-supervised learning for training interest point detectors and descriptors. Expanding upon SuperPoint, SuperGlue
\cite{sarlin2020superglue}
introduced a graph neural network that matches two sets of local features by jointly finding correspondences and rejecting non-matchable points. For the tasks such as feature matching and place recognition , both SuperPoint and SuperGlue have received widespread adoption \cite{keetha2021hierarchical}.

Some image retreival methods \cite{hausler2021patch} directly extract CNN-based image features. \cite{sharif2014cnn} produces a global image representation by aggregating the activation CNN features.
NetVLAD \cite{arandjelovic2016netvlad} uses a generalized end-to-end deep learning-based Vector of Locally Aggregated Descriptors (VLAD) \cite{jegou2010aggregating} layer. However, one of the main challenges faced by NetVLAD and other similar methods is the limited availability of training data, which can adversely affect performance.
To overcome this issue spatial/depth data has been incorporated 
\cite{zaki2019viewpoint}
and input modalities such as RGB-D images and point cloud data have been explored.

These descriptors are capable of producing distinguishable descriptions, but struggle in visually similar environments. In these conditions, different scenes could have similar local textures, which results in similar descriptions and finally leads to the failure of matching.

\subsection{Object semantic features and their application}

Object based semantic features are more robust and informative, and have been widely used in robotics applications such as SLAM. The pioneering work of SLAM++ \cite{salas2013slam++}  performs object-level SLAM using a depth camera.
\cite{qian2021semantic} develop a quadratic-programming-based semantic object initialization scheme to achieve high-accuracy object-level data association and real-time semantic mapping.
\cite{zhang2018semantic} integrated object detection and localization module together to obtain the semantic maps of the environment and improve localization. X-View \cite{gawel2018x} localize aerial-to-ground globally and ground-to-ground robot data of drastically different viewpoints using object graph descriptors based on random walks.

Recently, AirCode \cite{xu2022aircode} proposed a feature sparse and object dense encoding method which is robust to viewpoint changes, scaling, occlusion, and even object deformation. Building upon that, AirObject \cite{keetha2022airobject} introduced a temporal CNN across structural information in multiple frames to perform temporal 3D object encoding. These  frames were obtained from a graph attention based encoder.
However, using these object descriptors for relocalization still remains an open question. Taking motivation from above examples, 
we use object encoders, like AirCode, to extract object embeddings for relocalization.

\begin{figure}[th]
    \centering
    \includegraphics[width=\linewidth]{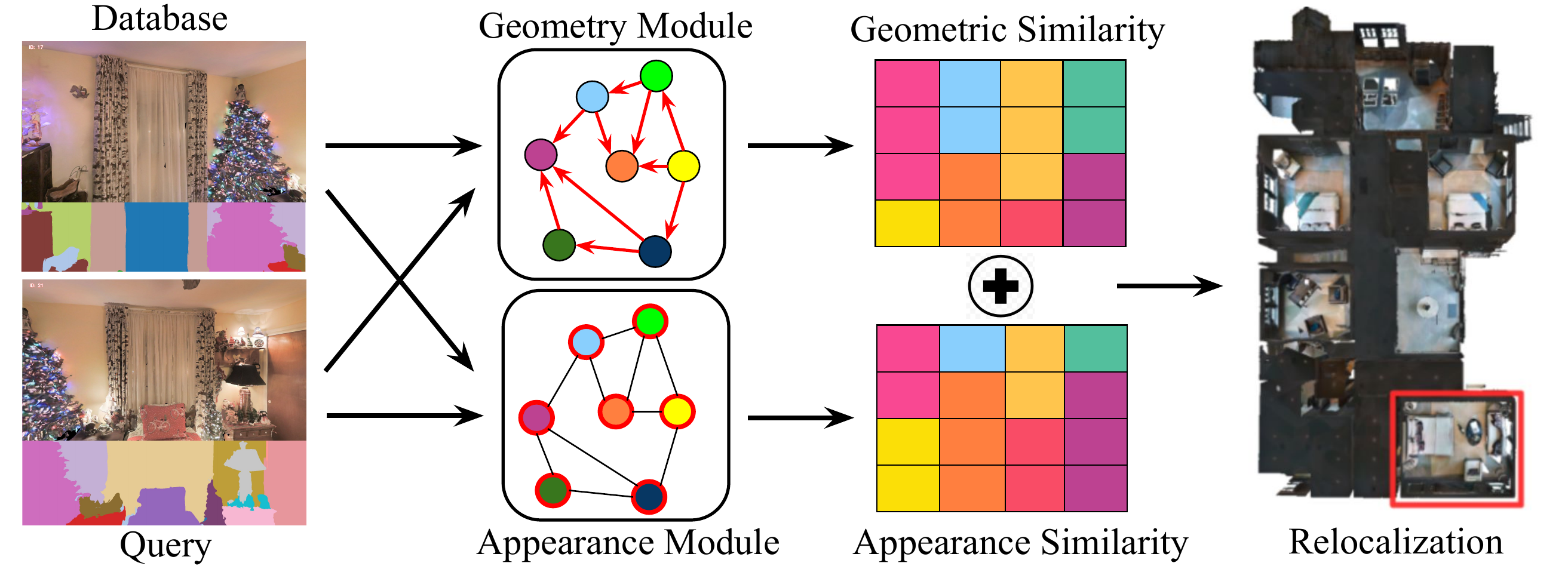}    \caption{
    The proposed object matching framework uses a geometry module and an appearance module to match query images with database objects for indoor relocalization.}
    \label{fig:Methodology}
\end{figure}

\section{Proposed Approach}

We propose AirLoc, a new architecture shown in \fref{fig:Methodology}.
It consists of two parts namely geometry and appearance module. In this section, we will first present the individual modules and then explain their ensembling. Finally, we will present the loss function for the geometry module.

\subsection{Appearance Module}

Appearance module encodes objects' visual characteristics. Typically, each room in our database has K ($K = 1,2,5,10$) images and a query consists of 1 to 2 images. Objects are first encoded into a feature vector, and if objects appear in more than one image, we take the arithmetic mean of their embeddings.
We then construct the database consisting room-wise object embeddings for relocalization.

\subsubsection{Object Encoders}

Instead of using masks or rectangular patches of objects, we extract their features using a group of key-points on the object, which can be more distinctive.
Based on previous research \cite{tarr2017concurrent}, we believe these key-points can provide robust object re-identification and can thus be used for embedding.
Specifically, we use Superpoint \cite{detone2018superpoint} to extract the feature points, where the position of each point is denoted as $\mathbf{h}_i = (x, y), i \in [1, N ]$, and the associated descriptor as $\mathbf{d}_i \in \mathbb{R}^{D_p} $, where $D_p$ is the descriptor dimension. 
We then group the points into objects using instance segmentations masks, which can be obtained from commonly-used networks like Mask R-CNN \cite{he2017mask} or an open-world object detector\cite{joseph2021towards}.

\begin{figure}[t]
    \centering
    \includegraphics[width=\linewidth]{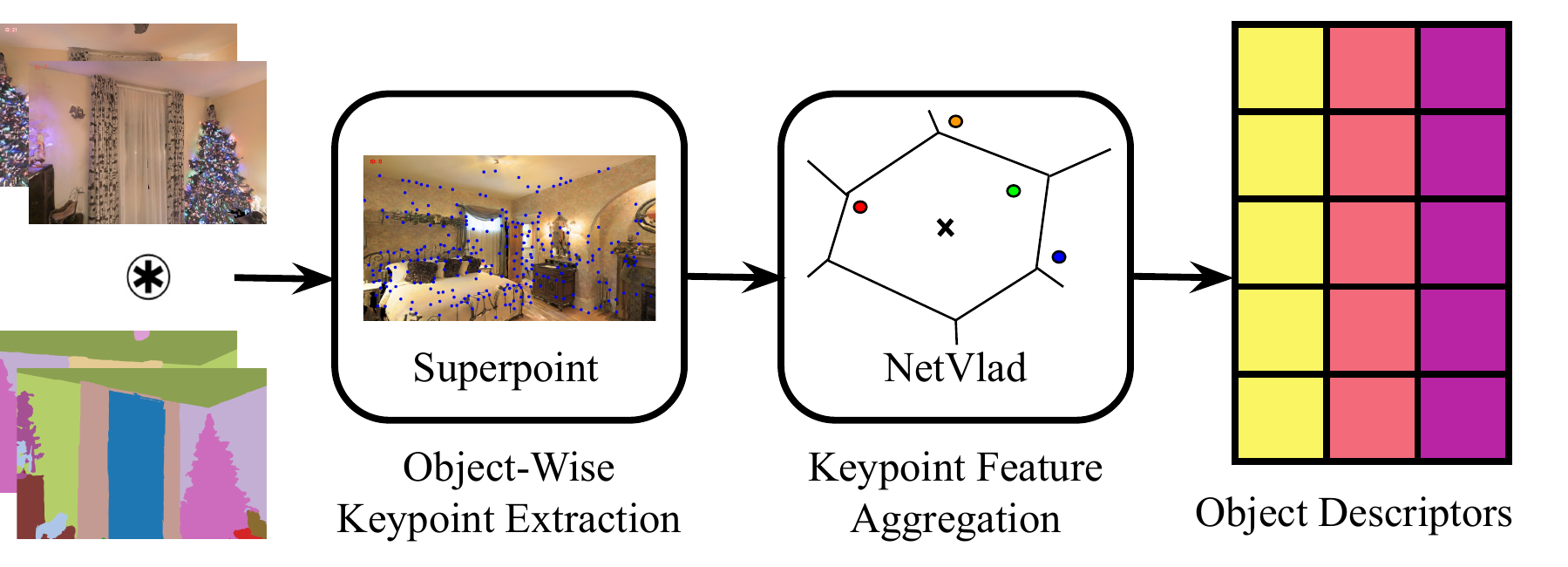}
    \caption{
    An object encoder is used in the appearance module to generate object descriptors for objects in a room using images and semantic labels.    
    }
    \label{fig:Database}
\end{figure}

Given the grouped points, we next aggregate the individual features to form a collective object encoding.
One of the most intuitive solutions is to use the graph-based networks such as GCN \cite{kipf2016semi} and GAT \cite{velivckovic2017graph} for feature aggregation, where each feature point is taken as a node.
However, we found that graph networks perform well when training and testing data are from the same distribution but can easily overfit to unseen environments.
On the contrary, image-based feature aggregation methods have a better generalization ability in this task.
For efficiency, we introduce a widely-used image retrieval framework NetVLAD \cite{arandjelovic2016netvlad} and modify it to fit our feature-point based representation as shown in \fref{fig:Database}. 
In the experiments, we found that this new framework can generalize to a new dataset, Reloc110 even if the our model is only trained on COCO~\cite{lin2014microsoft} and YT-VIS~\cite{yang2019video}, indicating its robustness to environmental changes. 
Given $N$ descriptors $\mathbf{d}_i$, $(i=1,\cdots,N)$, the object encoding $\mathbf{O}$ can be represented as a $C \times D_p $ dimensional vector:
\begin{equation}
\mathbf{O}(c) = \phi \left( \sum_{i=1}^{N}a_c(\mathbf{d}_i)(\mathbf{d}_i-\mathbf{x}_c)\right),
\end{equation}
where $\mathbf{O}(c)\in\mathbb{R}^{D_p}$ is the $c$-th row of $\mathbf{O}$. $\mathbf{x}_c$ is $c$-th cluster center ($c=1,\cdots,C$ and $C$ is predefined) and $a_c(\cdot)$ is the learnable parameter that denotes the soft assignment of descriptor $\mathbf{d}_i$ to cluster $\mathbf{x}_c$, and $\phi$ is a composed normalization function, i.e., an intra-normalization to make the model scale insensitive, followed by a L2-normalization to make the rows horizontally stacked into a vector.

\subsubsection{Similarity}
We propose an architecture to match the query with the database. Once the object descriptors are generated, they are exhaustively matched with the database using cosine similarity.
This results in an object similarity matrix $\mathbf{S}$, where each column consists similarity scores of a query object with all the candidate objects in the database.
This can be represented as:
\begin{equation}
\mathbf{S}(j,k) = \cos(\mathbf{O}_d(j), \mathbf{O}_q(k)),
\end{equation}
where $j$ and $k$ are $j$-th database object and $k$-th query object, $\cos$ is the cosine similarity, and $\mathbf{O}_d$ and $\mathbf{O}_q$ are database and query object embeddings, respectively.

\begin{figure}[t]
    \centering
    \includegraphics[width=\linewidth]{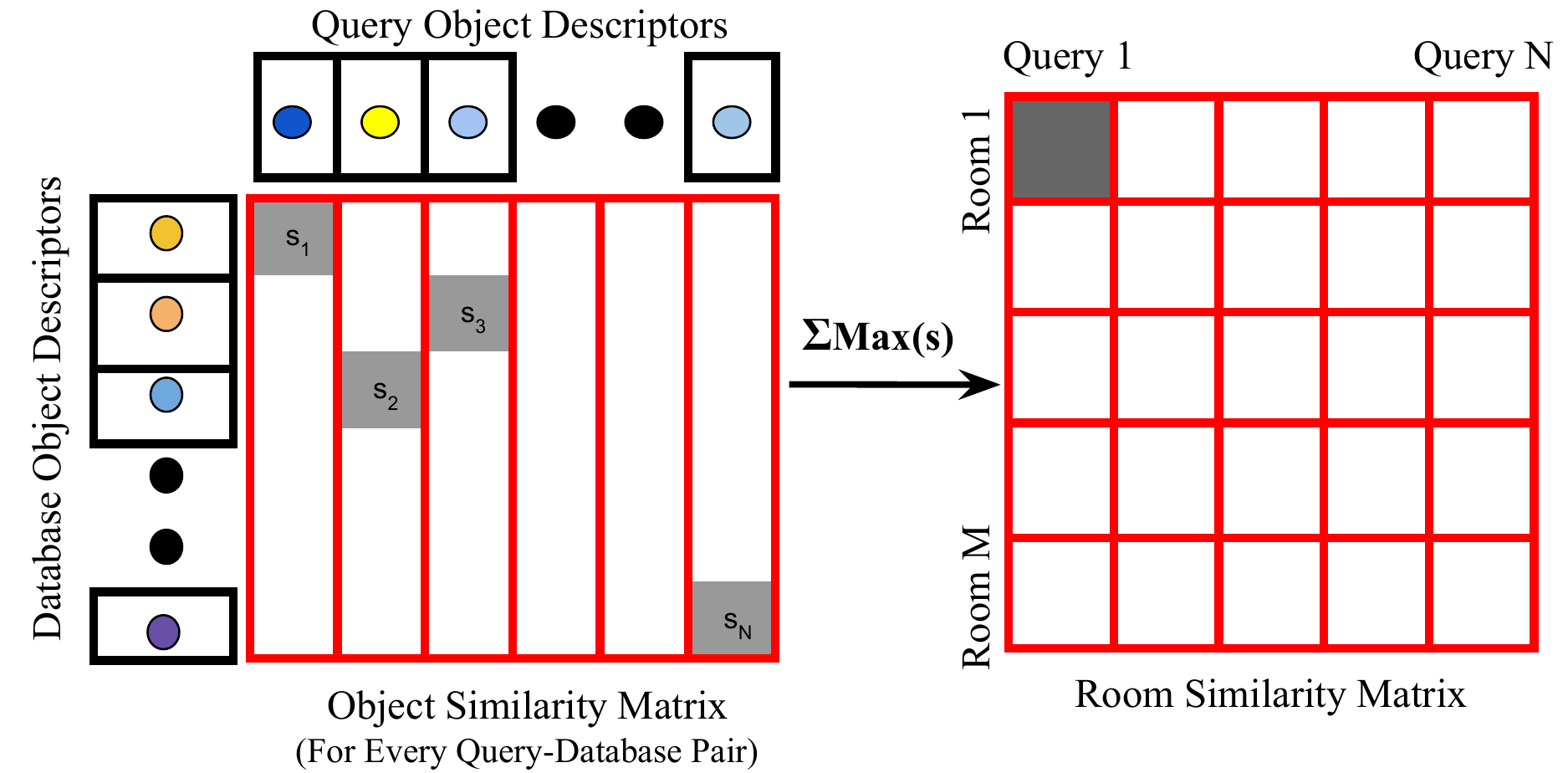}
    \caption{Appearance-based Matching: Maximum Object Similairty for every query-databse pair is summed up to form room similarity which is then used for relocalization.}
    \label{fig:Appearance}
\end{figure}

For efficiency, we adopt a simple yet effective object- and room-level matching framework, respectively, which is shown in \fref{fig:Appearance}.
The object-level matching is obtained by taking the maximum similarity with the database objects, while the room-level matching is obtained by summing up the object matching scores over each room.
This is because the matched rooms often have similar objects, and thus the summation of object similarities can reason about the room similarities, which can be represented as
\begin{equation}
\mathbf{R}(p,q) = \sum_{k=0}^{Z} \max(\mathbf{S}_{pq}(j,\mathrm{k})),
\end{equation}
where $\mathbf{R}$ is room similarity matrix and $\mathbf{S}_{pq}$ is object similarity matrix, $p$ belongs to database rooms and $q$ is the query room, and $Z$ is total number of query objects respectively.

\subsection{Geometry Module}

\begin{figure}[t]
    \centering
    \includegraphics[width=\linewidth]{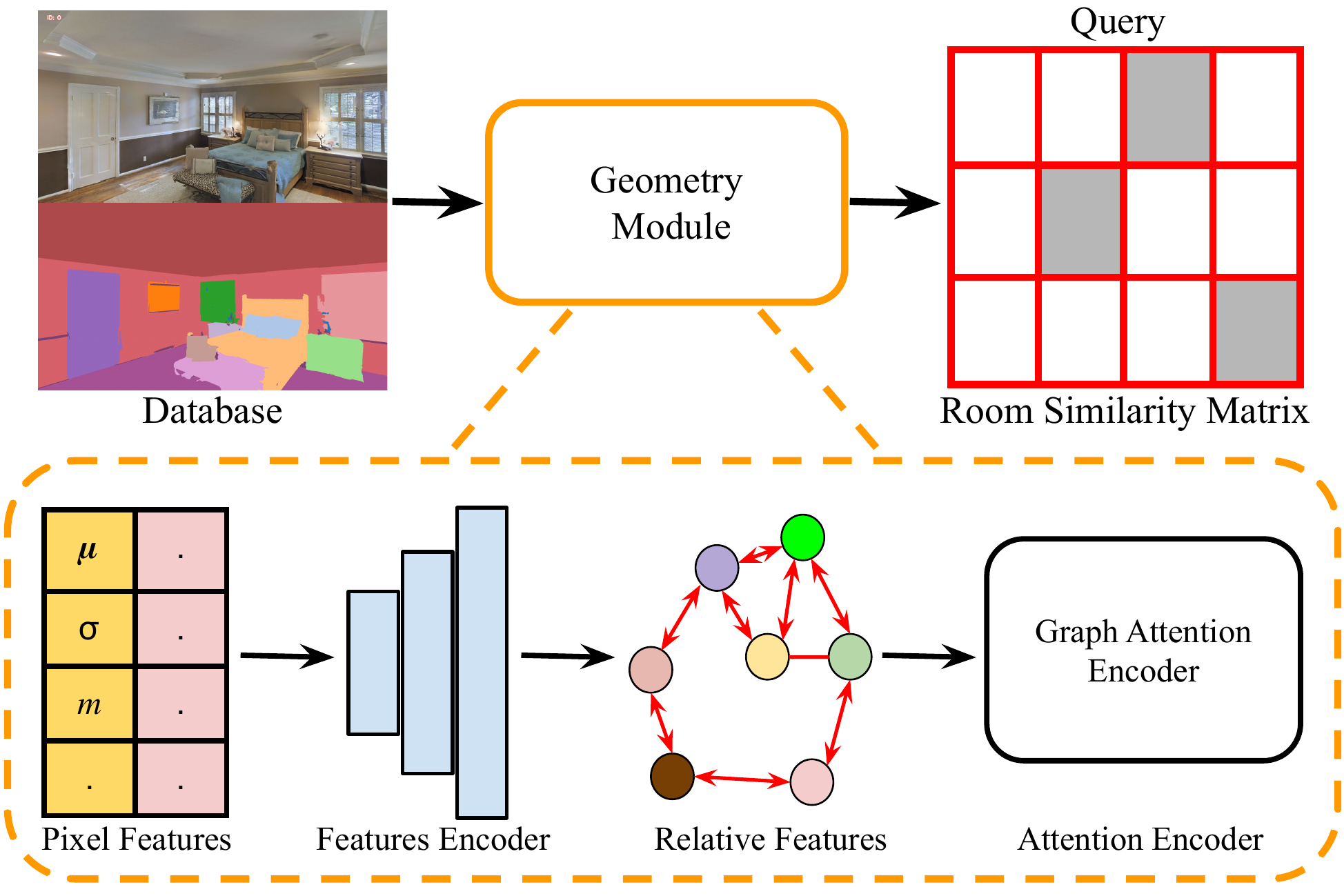}
    \caption{The structure of geometry module.}
    \label{fig:Geometry}
\end{figure}

Merely relying on appearance embedding has a potential problem, since rooms sharing  similar objects could be confusing. Inspired by the fact that objects are usually placed in different relative locations, we design a geometry module as shown in \fref{fig:Geometry} to assist the appearance-based matching.

An intuitive way to compute relative locations is to use depth measurements, but this makes the framework incompatible for cell-phone applications where depth information is often unavailable.
For better generalizability, we resort to object-wise key-point locations to encode geometric information.
Specifically, we use their mean location ($\mu_j$), standard deviation ($\sigma_j$), \nth{1}-, \nth{2}-, and \nth{3}-order momentum ($m_j^1,m_j^2,m_j^3$), and singular value decomposition ($\mathrm{svd}_j$).
Similar to appearance module,  if an object
appears in more than one image, we take the arithmetic mean of its geometric features.
Afterwards, the geometric features are passed through a multilayer perceptron (MLP) and then subtracted from each other to get relative geometric features.
In this way, if there are $Z$
objects we get $C_2^Z$ relative geometric features, which can be computed as:
\begin{align}
\mathbf{o}_j & = [\mu_j, \sigma_j, m_j^1, m_j^2, m_j^3, \text{svd}_j], \\
\mathbf{e}_{jk} & = g(\mathbf{o}_j) - g(\mathbf{o}_k),
\end{align}
where $[\cdot]$ is concatenation, $\mathbf{e}_{jk}$ is the relative location feature between $j$-th and $k$-th object and $g(\cdot)$ denotes MLP layer.

These geometric features are then passed through a two-layered GAT \cite{velivckovic2017graph} to 
perform attention-based message propagation between the location features
\begin{align}
\mathbf{e}_u^t & = \sigma \left(\sum_{v \in \mathcal{N}(u)} a_u \cdot \mathbf{W} \cdot \mathbf{e}_u^{t-1}\right), \\
\mathbf{r}  & =  \frac{\sum_{u=0}^U \mathbf{e}_u}{U},  
\end{align}
where $\mathbf{e}_u^t$ is the $u$-th location feature at $t$-th graph layer, $\sigma$ is nonlinearity, $a_u$ is attention coefficient, $\mathbf{W}$ is learnable weight matrix \cite{velivckovic2017graph} and $\mathbf{r}$ is room level embedding of dimension $E_o$.
Finally, cosine similarity matching of query and database room embeddings yields a room similarity matrix similar to the appearance module.

\begin{equation}
\mathbf{R}_{\text{loc}}(p,q) = \cos(\mathbf{r}_p, \mathbf{r} _q),
\end{equation}
where $\mathbf{r}_p$ and $\mathbf{r}_q$ are $p$-th database and $q$-th query room.

\subsection{Feature Ensembling}
After we get a set of room similarities based on appearance and geometry features, the final step is to integrate them using a weighted sum with the weight $w$. 
\begin{equation}
\mathbf{R'} = w \cdot \mathbf{R} + \mathbf{R}_\text{{loc}}
\end{equation}
Furthermore, it is observed that in most true positives from appearance-only matching, the similarity of the matched room is very high as compared to others. Hence, in such cases, there is not much need to use both modules. Therefore, in order to reduce the runtime and avoid the possibility of result degradation due to rooms having similar geometry but different objects, we apply the geometry-based assistance only to those queries where the difference between the appearance similarity of the highest and second-highest match is less than some threshold that we call the ``appearance threshold" ($T_{\text{diff}}$). The queries with a difference greater than the threshold are classified solely by appearance matching.

\begin{figure*}[!tb]
    \centering
    % \vspace{-5pt}
    \subfloat
    { \includegraphics[width=0.23\textwidth]{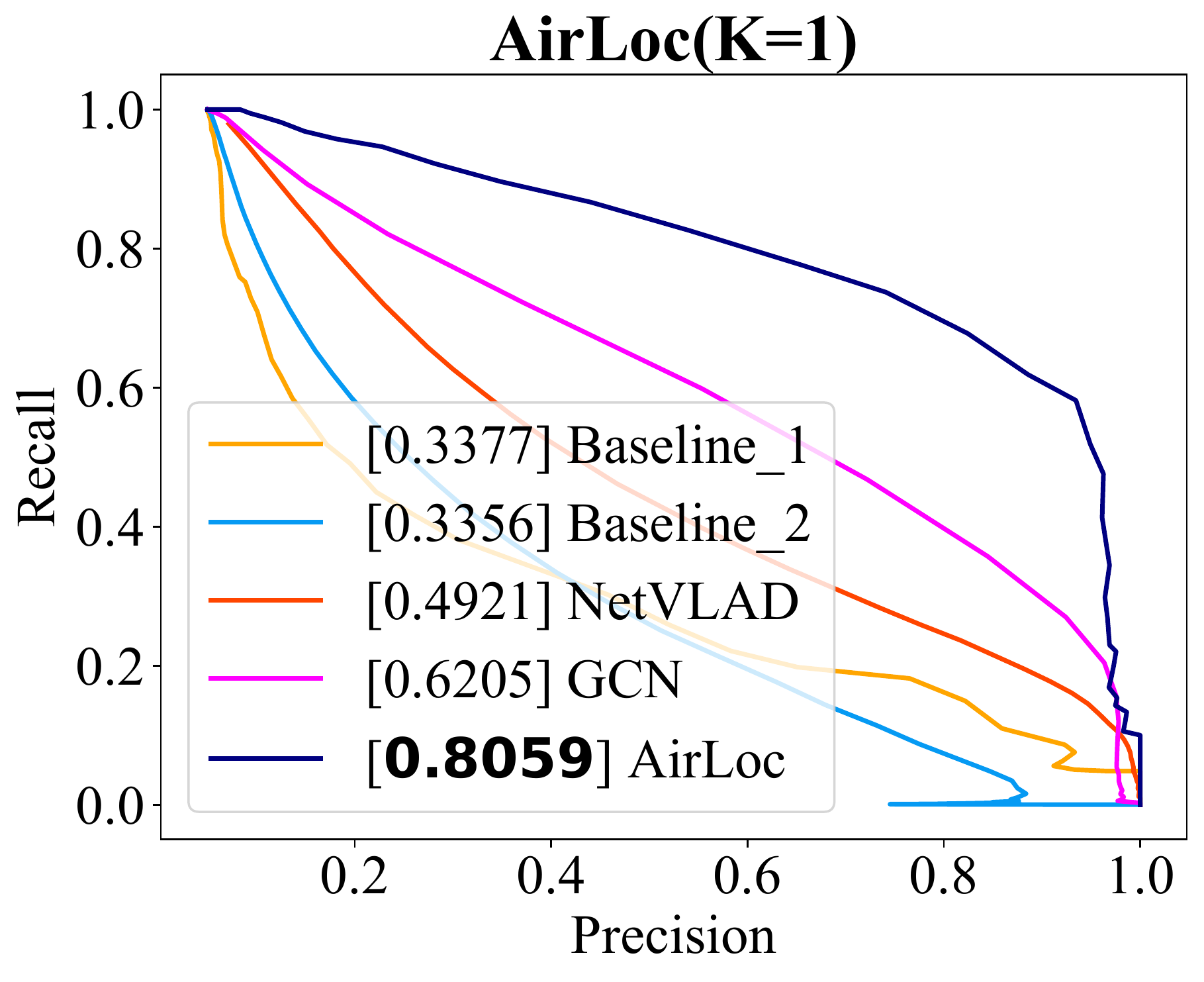}
        \label{fig:AirLoc_1} }
    \subfloat
    {\includegraphics[width=0.23\textwidth]{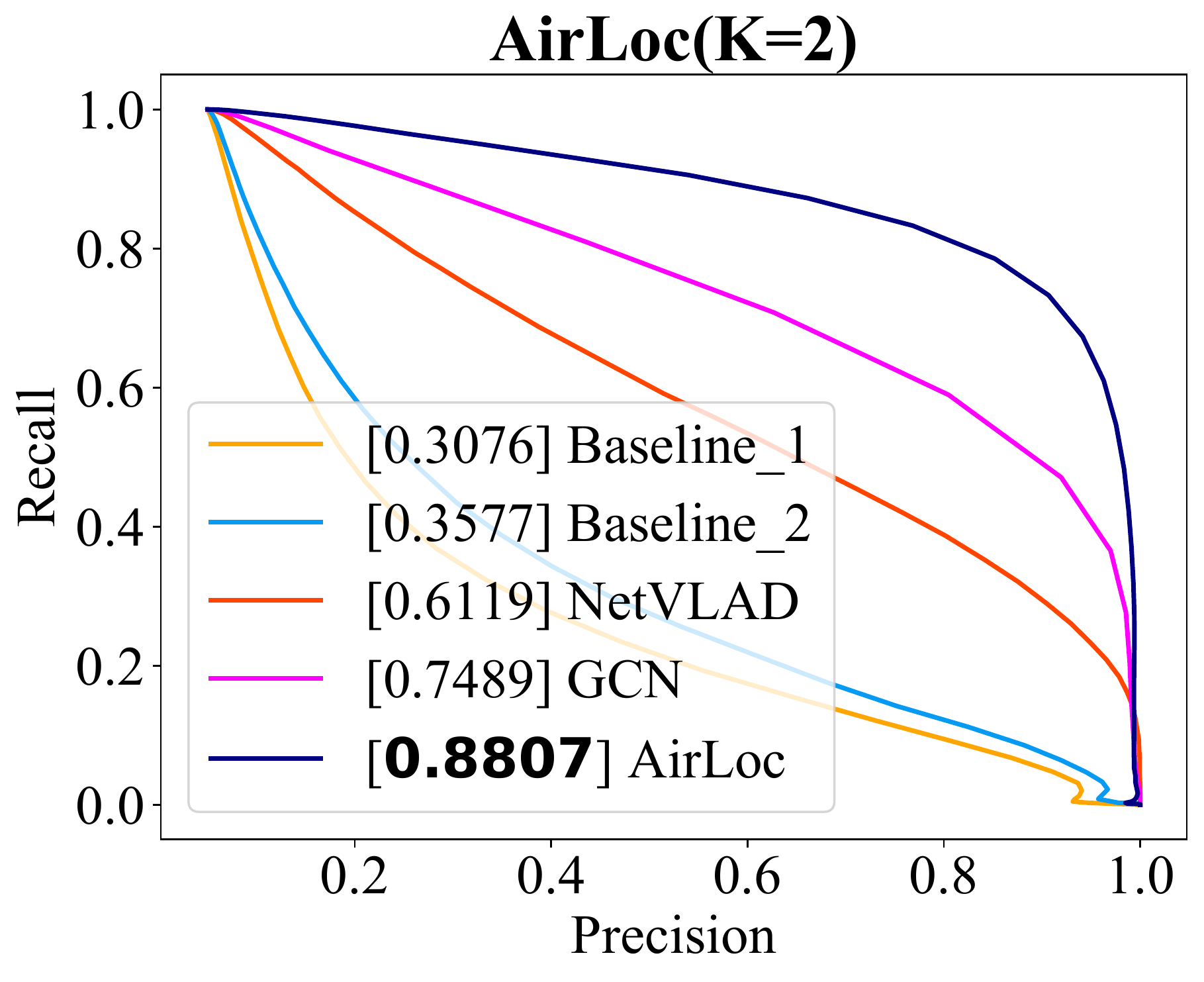}
        \label{fig:AirLoc_2}
    }
    \subfloat
    {\includegraphics[width=0.23\textwidth]{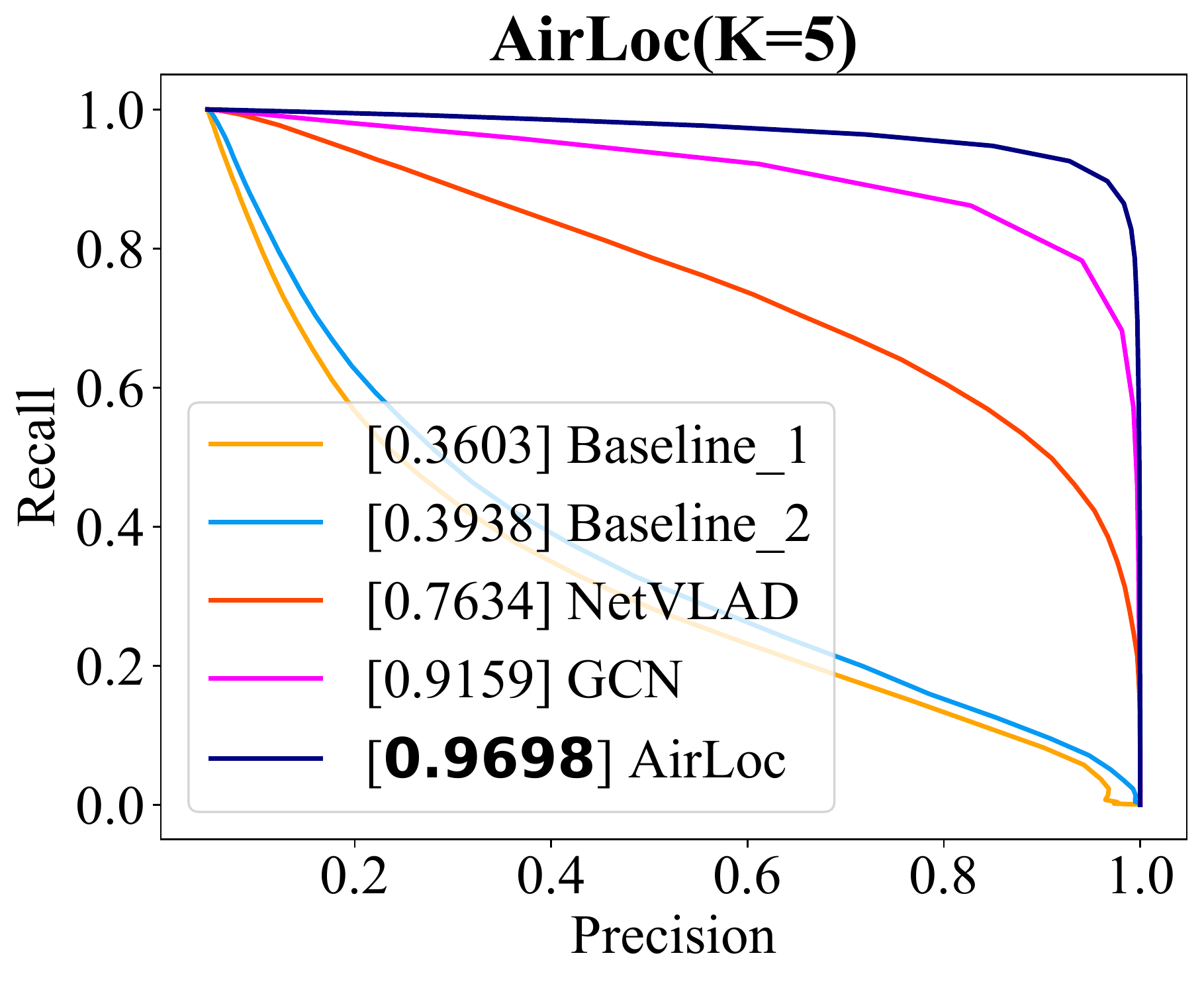}
        \label{fig:AirLoc_5}
    }
    \subfloat
    {\includegraphics[width=0.23\textwidth]{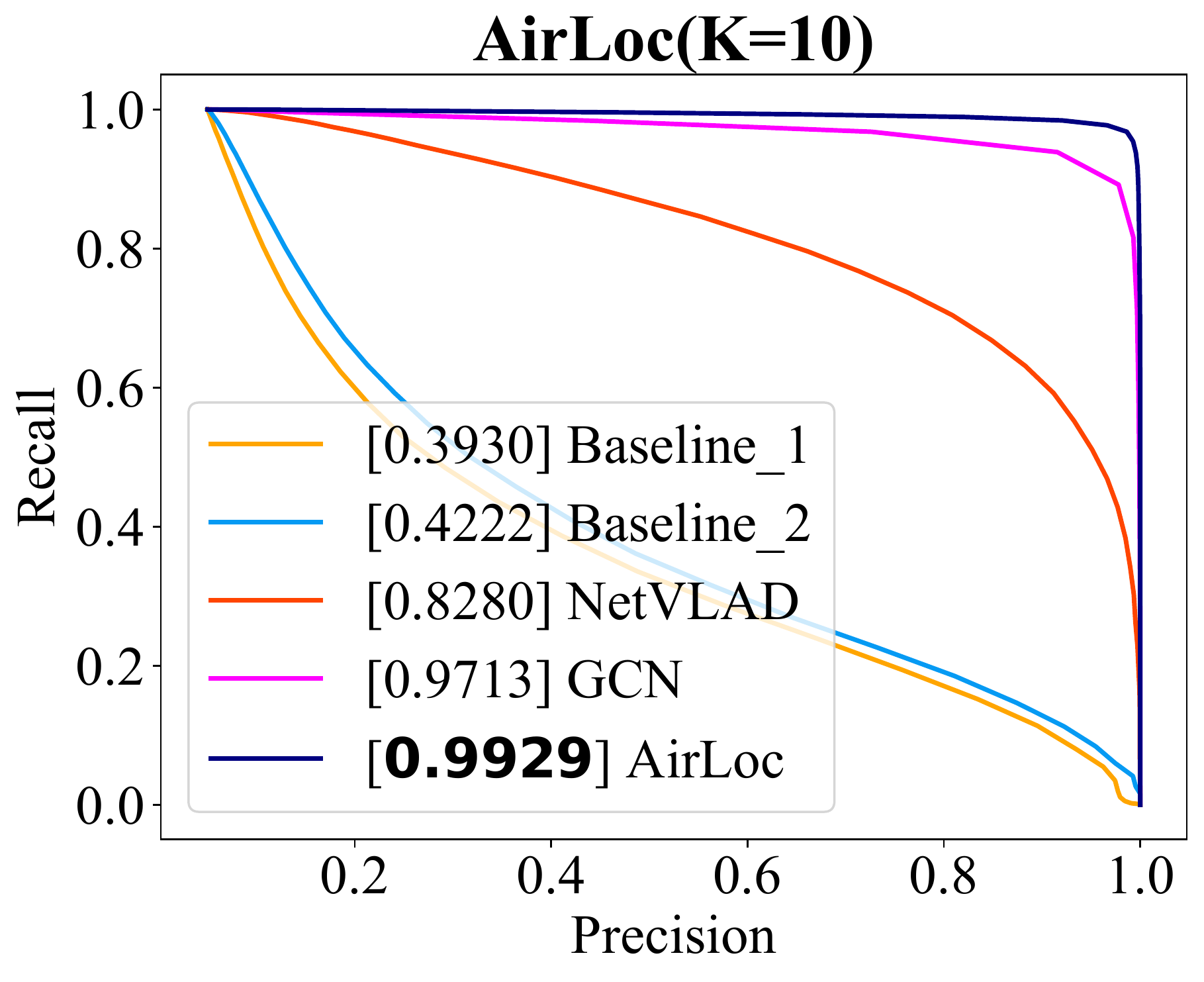}
        \label{fig:AirLoc_10}
    }
    \caption{Precision-Recall plots showing comparison between AirLoc and different baselines for different K values.}
    \label{fig:Results}
\end{figure*}

\subsection{Loss Function}
The graph attention encoder in geometry module is supervised by the room matching loss. The room matching loss $L_r$ maximizes the cosine similarity of positive room pairs and minimises the cosine similarity of negative room pairs.
\begin{equation}
\begin{aligned}
    L_r & = \sum_{\{p,q\} \in P^+} (1 - \mathrm{Cos}(\mathbf{r}_p, \mathbf{r}_q))\\
        & + \sum_{\{p,q\} \in P^-} \max(0, \mathrm{Cos}(\mathbf{r}_p, \mathbf{r}_q) - \zeta),
\end{aligned}
\end{equation}
where $\zeta = 0.2$ is a constant margin, $\mathrm{Cos}$ is the cosine similarity, and $P^+, P^-$ are positive and negative object pairs.

\section{Experimental Results}
\subsection{Dataset}
The dataset adopted in this work, named Reloc110, is newly rendered using Habitat-Sim \cite{savva2019habitat}, which is a high-performance physics-enabled 3D simulator supporting 3D scans of indoor/outdoor spaces and rigid-body mechanics. 
To minimize the gap between simulation and real-world, we borrowed Matterport3D \cite{chang2017matterport3d}, a large-scale RGB-D dataset that contains 90 building-scale scenes. 
All the Matterport3D scenes are in the form of textured 3D meshes and are created from real-world RGB-D images.

\begin{table}[t]
    \caption{\fix{Statistics of the newly rendered Reloc110 dataset. We present the names, images, and rooms of 15 scenes.}}
    \label{tab: Dataset_statistics}
    \centering
\begin{tabular}{p{0.15\linewidth}p{0.1\linewidth}p{0.1\linewidth}p{0.15\linewidth}p{0.1\linewidth}p{0.1\linewidth}}
\toprule[1.5pt]
Scene                             & Images                     & Rooms                  & Scene                            & Images                           & Rooms                         \\ 

\midrule
8WUm & 18803 & 8 & ULsK & 13600       & 5                             \\  
EDJb                       & 22800                      & 8                      & Vzqf                      & 27000                            & 9                             \\  
i5no                       & 18200                      & 7                      & wc2J                      & 32800                            & 12                            \\  
jh4f                       & 13400                      & 5                      & WYY7                      & 11200                            & 5                             \\  
mJXq                      & 25199                      & 9                      & X7Hy                      & 17800                            & 7                             \\  
qoiz                      & 25500                      & 9                      & YFuZ                  & 20800              & 8                             \\  
RPmz                       & 17800                      & 6                      & yqst                      & 15600                            & 6                             \\ 
S9hN                     & 25000                      & 9                      & \textbf{Total}  & \textbf{306000} & \textbf{113} \\ \bottomrule[1.5pt]
\end{tabular}
\end{table}

We selected 15 scenes from the dataset, each containing approximate 8 rooms. 
For every room, we sample approximate 2500 random poses, which are easily accessible for a human or a robot, i.e., not inside a wall or under the ground. 
Therefore, the images corresponding to the poses are similar to what humans or robots perceive in their general actions. 
We then render corresponding RGB image and semantic segmentation labels for all the collected poses. 
The dataset contains a total of 306000 images divided into 113 rooms. 
 \tref{tab: Dataset_statistics} shows total number of rooms and total images generated for every scene. 
 We further divide the dataset into test and train split as well where 3 scenes (RPmz, S9hN, ULsK) are test split and remaining are train split.

\subsection{Implementation Details}\label{sec:implementation}
The AirLoc configurations for appearance based matching are superpoint input dimension $D_p = 256$ and the number of clusters in NetVLAD is $C= 32$. 
Configuration for geometric matching are:  relative location feature dimension $E  = 256$, hidden dimension of graph layer $E_h = 512$, output dimension of graph $E_o = 1024$. For GAT we use 8 heads and dropout of 0.5.  For training, we used a batch size of 256, learning rate of $1e^{-4}$. The network is trained for 30 epochs using Adam optimizer on a Nvidia A100 80GB GPU.
 
To validate the generalizability of the AirLoc, we do not train appearance module on Reloc110 dataset. 
Instead, we use NetVLAD pretrained on COCO~\cite{lin2014microsoft} and YT-VIS~\cite{yang2019video} datasets. 
The train split is only used for learning the geometry module which only considers the relative position of objects and hence can easily be generalized to unseen rooms.

For evaluation of room level localization performance, we use the test split of Reloc110 dataset. To switch between appearance-only and appearance-geometry matching, we use ($T_{\text{diff}}$) as 0.1. The weight ($w$) for the weighted sum of appearance and geometry is 10. 

\subsection{Evaluation Metrics}
AirLoc's performance is evaluated on two metrics, accuracy and precision-recall. While computing accuracy, we make a one-one matching where a query is matched only with one most similar room. Accuracy is then calculated as ratio of correctly matched queries to total number of queries.
However, while computing precision-recall, we allow a one to many matching. The query-database pair having similarity value higher than a threshold $\rho$ is considered as a match. Based on True Positives and False Negatives, we  calculate precision and recall. Furthermore, by varying the threshold values $\rho \in (0,1)$, we obtain precision-recall curves and calculate area under curves (AUCs).

\begin{table}[t]
    \caption{\fix{The Results Comparing AirLoc with baselines}.}
    \label{tab:results_table}
    \centering
    \begin{tabular}{p{0.2\linewidth}p{0.1\linewidth}p{0.1\linewidth}p{0.1\linewidth}p{0.1\linewidth}p{0.1\linewidth}}
        \toprule[1.5pt]
        \multirow{2}{*}{Method} & \multicolumn{5}{c}{Accuracy} \\
 &K=1 & K=2 & K=3 & K=5 & K = 10 \\
        \midrule
        Baseline\_1 & 40.64 & 61.55 & 69.88 & 78.69 & 84.81\\
        Baseline\_2 & 41.31 & 49.14 & 51.65 & 61.67 & 65.64 \\
        NetVLAD \cite{arandjelovic2016netvlad} & 58.01 & 74.16  & 79.89 & 90.02 & 95.37\\
        GCN \cite{velivckovic2017graph} & 61.31 & 76.57 & 86.30 & 91.61 & 96.62\\
         AirLoc   & \textbf{75.35} &\textbf{87.26} &\textbf{91.75} &\textbf{94.35} &\textbf{98.32}\\ 
        \bottomrule[1.5pt]
    \end{tabular}
\end{table}

\begin{table*}[t]
    \caption{\fix{Precision-Recall Results Comparing AirLoc with baselines}.}
    \label{tab:pr_results_table}
    \centering
    % \begin{tabular}{C{0.25\linewidth}C{0.12\linewidth}C{0.12\linewidth}C{0.22\linewidth}}
    \begin{tabular}{p{0.07\linewidth}p{0.03\linewidth}p{0.03\linewidth}p{0.05\linewidth}p{0.03\linewidth}p{0.03\linewidth}p{0.05\linewidth}p{0.03\linewidth}p{0.03\linewidth}p{0.05\linewidth}p{0.03\linewidth}p{0.03\linewidth}p{0.05\linewidth}p{0.03\linewidth}p{0.03\linewidth}p{0.03\linewidth}}
        \toprule[1.5pt]
        \multirow{2}{*}{Method} & \multicolumn{3}{c}{K = 1}  & \multicolumn{3}{c}{K = 2} &  \multicolumn{3}{c}{K = 3} & \multicolumn{3}{c}{K = 5} & \multicolumn{3}{c}{K = 10} \\
        \cmidrule(r{10pt}){2-4}\cmidrule(r{10pt}){5-7}\cmidrule(r{10pt}){8-10}\cmidrule(r{10pt}){11-13}\cmidrule{14-16}
 & P & R & F-1  &
 P & R & F-1  &
 P & R & F-1  &
 P & R & F-1  &
 P & R & F-1  \\
        \midrule
        Baseline\_1 & 92.30 & 8.63 & 15.78 & 55.31 & 19.35 & 28.67 & 37.67 & 34.17 & 35.84 & 26.61 & 47.34 & 34.07 & 21.33 & 57.78 & 31.16\\
        Baseline\_2 & 73.18 & 11.38 & 19.71 & 53.34 & 25.55 & 34.55 & 39.26 & 35.67 & 37.38 & 31.93 & 46.42  & 37.83 & 27.36 & 54.85 & 36.51\\
        NetVLAD & 100 &  0.4 & 0.8  & 98.65 & 16.20 & 27.83 & 67.57 & 53.16 & 59.51 & 33.43 & 87.17 &48.33 & 16.22 & 97.93 & 27.84 \\
        GCN & 72.17 & 46.79 & 56.77 & 80.49 & 58.94 & 68.05 & 90.44 & 68.04 & 77.66 & 94.06 & 78.26 & 85.44 & 97.80 & 89.17 & 93.29 \\
         AirLoc   & 82.43 & 67.77 & \textbf{74.39} & 90.66 & 73.27 &\textbf{81.05} &  94.63 & 78.19 & \textbf{85.63} & 98.33 & 86.47 &\textbf{92.02} & 99.27 & 95.40 & \textbf{97.30}\\ 
        \bottomrule[1.5pt]
    \end{tabular}
\end{table*}

\subsection{Comparison to State-of-the-art Methods}

AirLoc is compared with two types of baselines: room-level and object-level. The room-level baselines: Baseline\_1, Baseline\_2, and NetVLAD, extract room-level features from the input and calculate a room similarity matrix, thereby avoiding object matching. In Baseline\_1, NetVLAD-based object encoders are used to extract individual objects features and the room features are then calculated by averaging the output object embeddings. In Baseline\_2, the object encoder from Baseline\_1 is replaced with GAT, allowing a comparison of the performance of NetVLAD and GAT for object encoding. The NetVLAD baseline uses the output image descriptors from a NetVLAD module as room features, similar to how NetVLAD is typically used for place recognition \cite{arandjelovic2016netvlad}. It is worth noting that in this baseline, the NetVLAD module is not used for object encoding, but rather for encoding the entire image.

The object-level baseline, GCN \cite{velivckovic2017graph}, extract object information first and matches object-level data to further generate room similarity scores. It uses a similar architecture as AirLoc, but with two differences. First, the NetVLAD-based object encoder used in AirLoc is replaced with a graph attention-based object encoder. This allows for a comparison of the performance of these two types of object encoders. Second, the geometry module is not used in the GCN baseline which means that the it does not incorporate information about the spatial relationships between objects.

In \fref{fig:Results} and \tref{tab:pr_results_table}, the performance of AirLoc is compared to the baseline methods using precision-recall and F-1 score, respectively, for different values of K. The results show that AirLoc consistently outperforms all the baselines across all K values in both PR-AUC and F-1. In particular, AirLoc exceeds GCN and NetVLAD  by an average of 9.5\%, 22.5\% respectively in terms of PR-AUC and 10\% and 49\% respectively in terms of F-1 score. It can also be noticed that for both the metrics, the performance gap between object-based methods and room-based methods is consistently large, demonstrating the importance of object level data.

\begin{figure}[t]
    \centering
    \includegraphics[width=\linewidth]{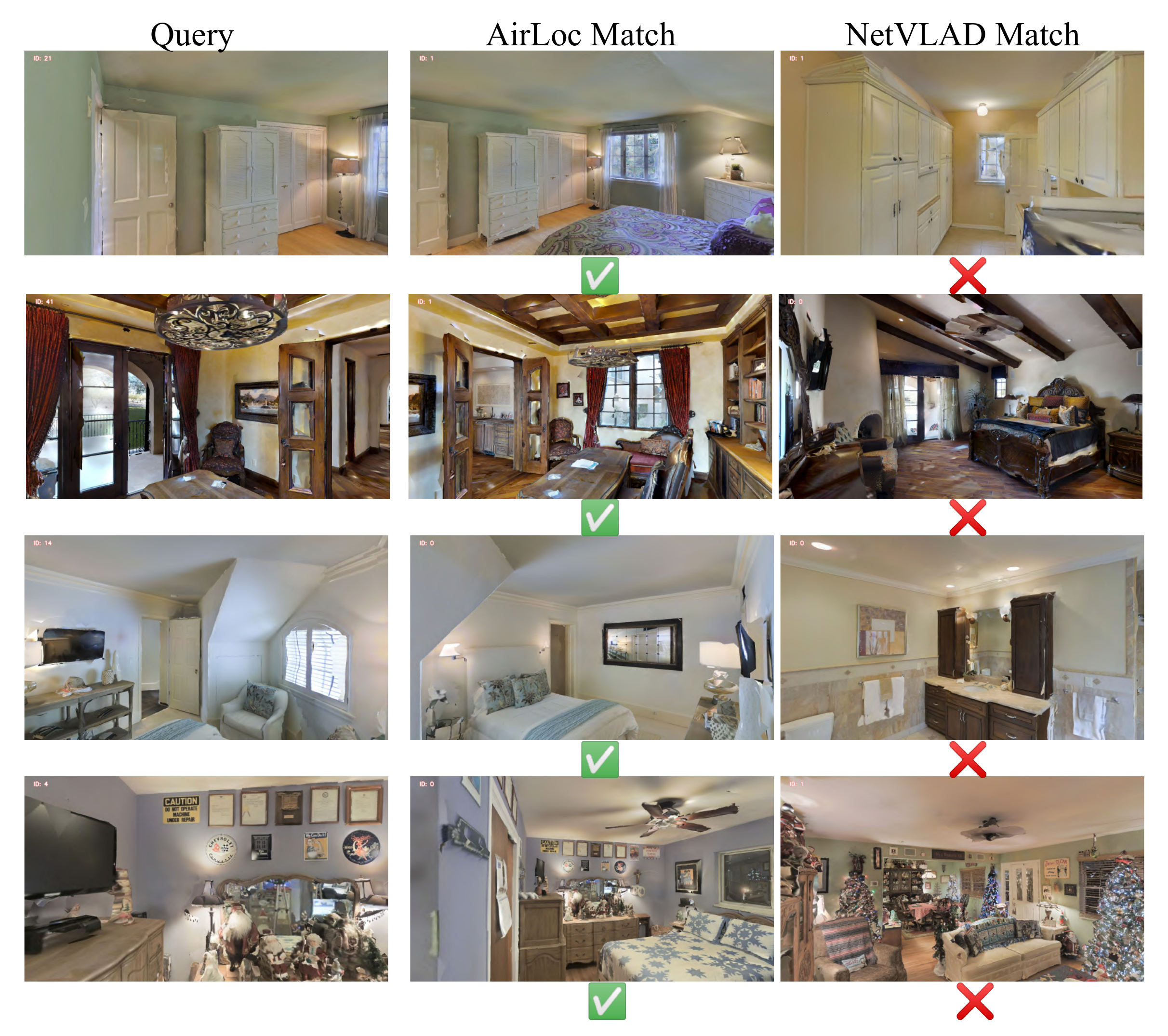}
    \caption{Qualitative Results.}
    \label{fig:qulatitative_results}
\end{figure}

\tref{tab:results_table} presents comparisons of AirLoc and baseline methods in terms of accuracy. AirLoc outperforms all other approaches in accuracy as well. Specifically, it outperforms GCN and NetVLAD by an average of 7\% and 10\%, respectively, and the margin of improvement is larger when K is smaller, indicating that AirLoc does not need as many images to perform well compared to the other methods.

In \fref{fig:qulatitative_results} we present examples demonstrating the difference in performance between NetVLAD and AirLoc. For a query, the closest database image produced by NetVLAD is shown in the right column, while the closest database image produced by AirLoc is shown in middle column. It can be observed that NetVLAD's matches look more visually similar to query, but objects in these images are different from query objects. This results in a wrong match for NetVLAD. In contrast, AirLoc relies on object level data and is able to correctly match the query image even though the two images do not look visually similar. This demonstrates the effectiveness of using object-level data, as opposed to relying solely on visual similarity.

\subsection{Efficiency}

\tref{tab:running} presents overall runtime and inference time of individual modules for AirLoc. The runtime of the geometry module, which does not run constantly and whose inference depends on the appearance threshold ($T_{\text{diff}}$), is 4.8ms, much lower than the appearance module. The overall running time of AirLoc is about 20.4ms, satisfying real-time requirements of most applications. Even though the GCN baseline does not have a geometry module, its overall runtime is higher than that of AirLoc. This is due to the longer time taken by the node encoding of GCN, which uses a GAT rather than NetVLAD.
The NetVLAD has a lower runtime compared to other methods as it encodes the entire image, rather than individual objects. However, the accuracy and PR-AUC for NetVLAD are much lower than those of AirLoc.

 \begin{table}[t]
\caption{Runtime Analysis.}
\label{tab:running}
\centering
\begin{tabular}{>{\centering\arraybackslash}p{0.12\linewidth}>{\centering\arraybackslash}p{0.21\linewidth}>{\centering\arraybackslash}p{0.14\linewidth}>{\centering\arraybackslash}p{0.14\linewidth}>{\centering\arraybackslash}p{0.14\linewidth}}
\toprule[1.5pt]
Module & Node Encoding & Appearance & Geometry & Overall \\
\midrule
AirLoc & 2.5 \milli\second & 13.1 \milli\second & 4.8 \milli\second & 20.4 \milli\second \\

GCN & 8.1  \milli\second & 14.3 \milli\second & - \milli\second & 22.4 \milli\second \\
   
\bottomrule[1.5pt]
\end{tabular}

\end{table}

\subsection{Ablation Studies}

To evaluate the effectiveness of the geometry module, we compare the performance of AirLoc with and without the geometry module, as well as with $T_{\text{diff}} = 1$. $T_{\text{diff}} = 1$ means that every query is evaluated using appearance-geometry matching, opposed to AirLoc where some queries were evaluated using appearance matching only. The results, shown in \tref{tab:ablation_table}, demonstrate that AirLoc outperforms AirLoc without the geometry by an average of 1.2\%. This suggests that geometry module helps the system to reason about the geometry of the scene, leading to better and more accurate relocalization. Additionally, except for $K=2$, the performance of AirLoc with $T_{\text{diff}} = 1$ is lower than that of AirLoc without geometry module, indicating that the current setting with $T_{\text{diff}} < 1$ can be generalized to most cases.

\begin{table}[t]
    \caption{\fix{Ablation studies.}}
    \label{tab:ablation_table}
    \centering
    \begin{tabular}{p{0.35\linewidth}p{0.13\linewidth}p{0.13\linewidth}p{0.13\linewidth}}
        \toprule[1.5pt]
         \multirow{2}{*}{Method} & \multicolumn{3}{c}{Accuracy} \\
 &K=1 & K=2 & K=3\\
 \midrule
AirLoc ($T_{\text{diff}}$ = 1)   & 73.27 & 87.20 & 89.58 \\
AirLoc (w/o Geometry)   & 74.14 & 85.86 & 90.97 \\
AirLoc   & \textbf{75.35} &\textbf{87.26} &\textbf{91.75} \\
        \bottomrule[1.5pt]
    \end{tabular}
\end{table}

\subsection{Parameter Analysis}

To study the effect of different hyperparameters on the accuracy of the AirLoc system, we conducted a parameter analysis by varying the values of the hyperparameters and measuring the resulting performance. The results of the analysis, shown in \tref{tab:appearance_table}, demonstrate that the maximum accuracy for most values of K occurs around $T_{\text{diff}} = 0.1$, leading us to choose this value for the appearance threshold. The results in \tref{tab:m_table} show that the accuracy is highest for a appearance to geometry weight value of $w = 10$. These results provide insights into the impact of different hyperparameter values on the accuracy of the AirLoc system.

\begin{table}[t]
    \caption{\fix{Variation of accuracy with} $T_{\text{diff}}$.}
    \label{tab:appearance_table}
    \centering
    \begin{tabular}{p{0.1\linewidth}p{0.1\linewidth}p{0.1\linewidth}p{0.1\linewidth}p{0.1\linewidth}p{0.1\linewidth}}
        \toprule[1.5pt]
        \multirow{2}{*}{$T_{\text{diff}}$} & \multicolumn{5}{c}{Accuracy} \\
 & K=1 & K=2 & K=3 & K=5 & K=10 \\
        \midrule
        0.01 & 74.34 & 86.02 & 91.36 & 94.04 & \textbf{98.45}\\
        0.05 & 75.12 & 86.80 & 91.60 & 94.14 & 98.42 \\
        0.1 & \textbf{75.35} & 87.26  & \textbf{91.75} & \textbf{94.35} & 98.32\\
        0.2 & 75.17 & \textbf{87.58} & 91.40 & 93.62 & 97.95\\
        0.35 & 74.59 &  87.46 & 90.62  & 92.78 & 97.44\\
        0.5 & 74.14 & 87.28 & 90.12 & 92.12 & 96.97\\
        \bottomrule[1.5pt]
    \end{tabular}
\end{table}

\begin{table}[t]
    \caption{\fix{Variation of accuracy with} $w$.}
    \label{tab:m_table}
    \centering
    \begin{tabular}{p{0.1\linewidth}p{0.1\linewidth}p{0.1\linewidth}p{0.1\linewidth}p{0.1\linewidth}p{0.1\linewidth}}
        \toprule[1.5pt]
        \multirow{2}{*}{$w$} & \multicolumn{5}{c}{Accuracy} \\
 &K=1 & K=2 & K=3 & K=5 & K = 10 \\
        \midrule
        1 & 72.38 & 85.14 & 89.28 & 92.68 & 97.71\\
        5 & 74.75 & 87.10 & 91.07 & 93.97 & 97.98\\
        10 & \textbf{75.35} & 87.26 &\textbf{91.75} & \textbf{94.35} & 98.32\\
        20 & 75.10 &  \textbf{87.36} & 91.36  & 94.31 & 98.15\\
        50 & 74.08 & 87.27  & 89.63 & 93.22 & \textbf{98.41}\\
        \bottomrule[1.5pt]
    \end{tabular}
\end{table}

\subsection{Real-World Demo}
\begin{figure}[!tb]
    \centering

    \subfloat[Illumination Changes.]
    {\includegraphics[width=0.98\linewidth]{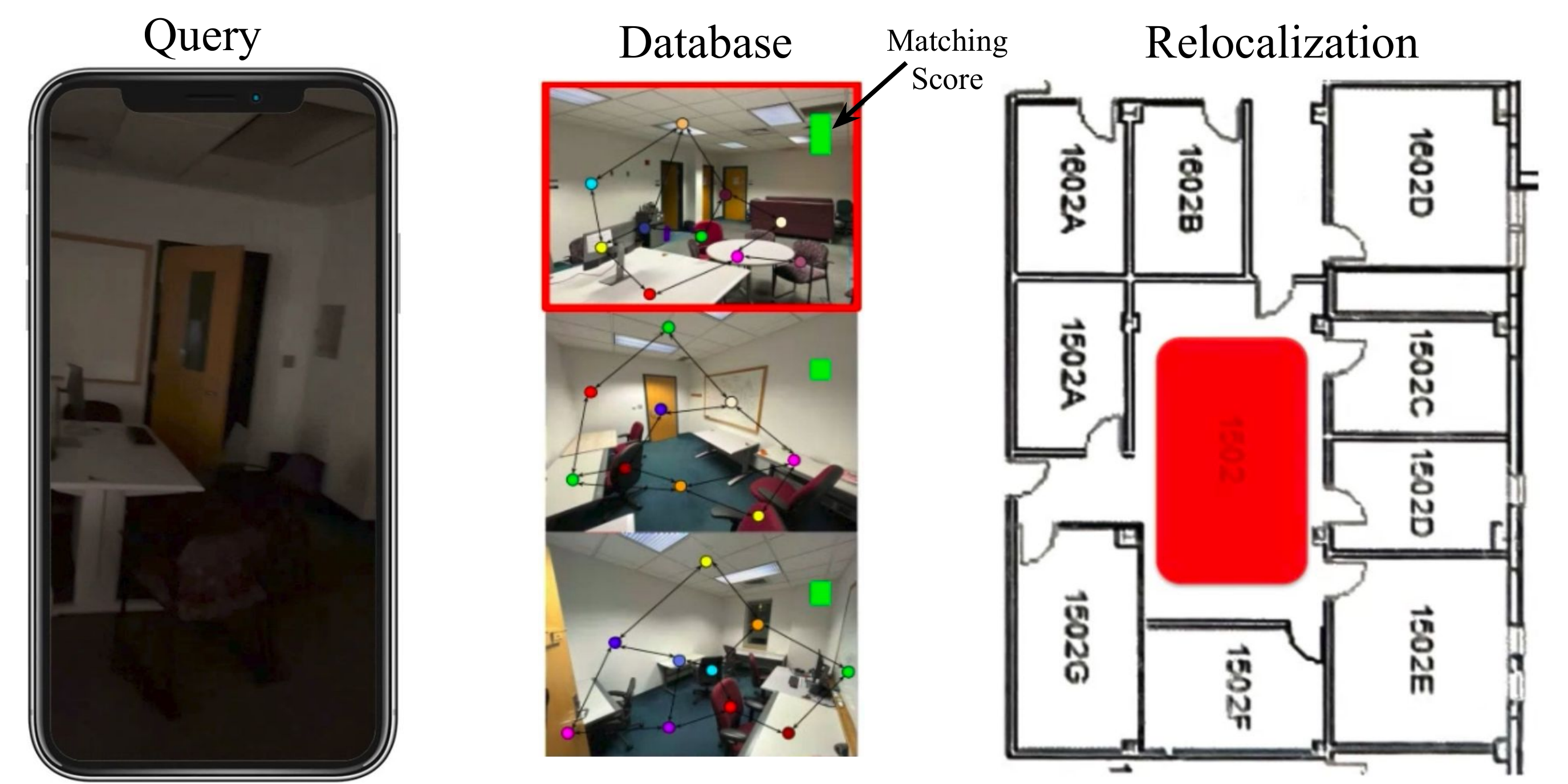}%
        \label{fig:illumination}
    }
    \hfil
    \subfloat[Human Interference.]
    {\includegraphics[width=0.98\linewidth]{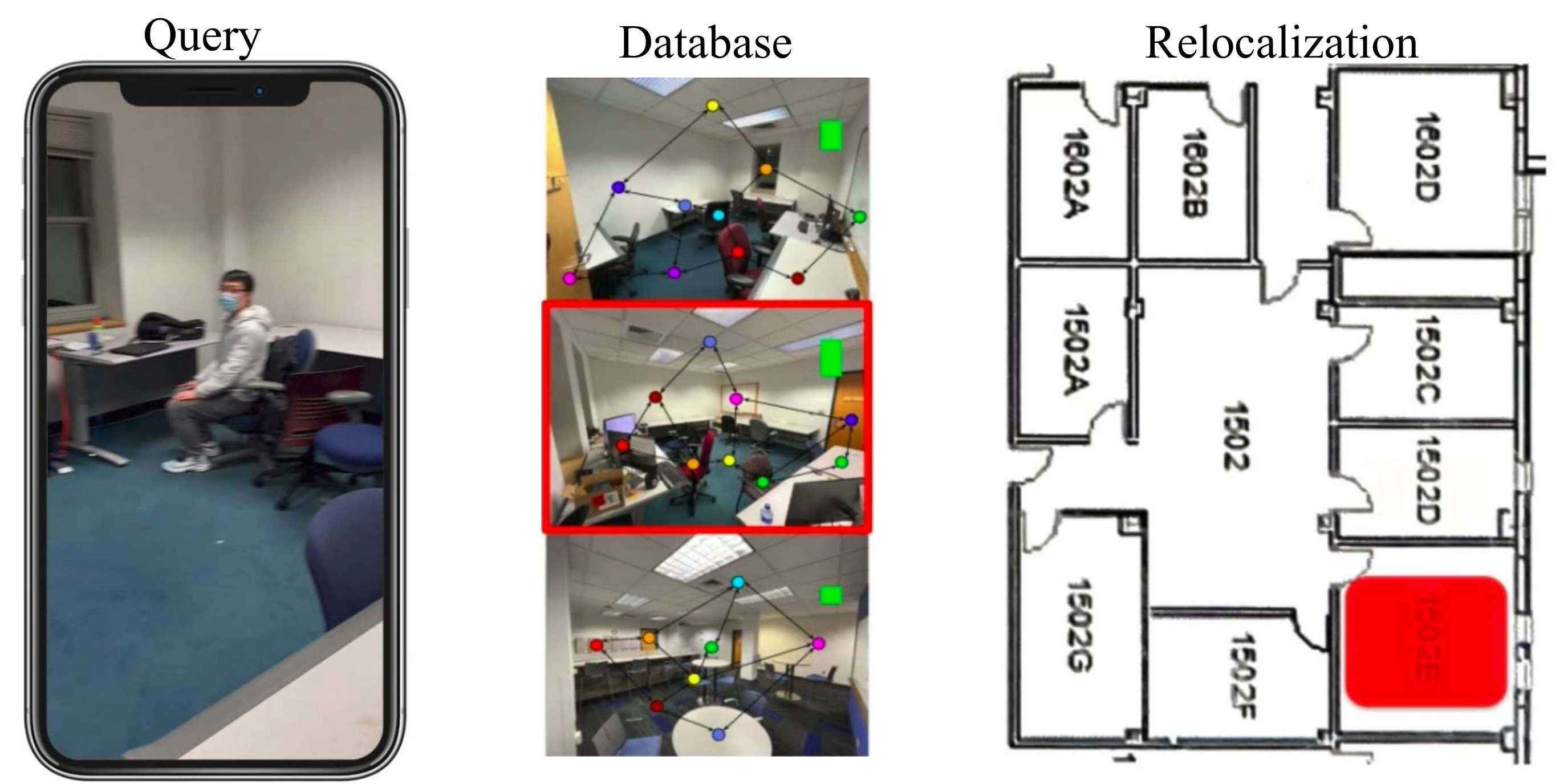}%
        \label{fig:human}
    }
    \caption{The live relocalization demo.}
    \label{fig:livedemo}
\end{figure}

This section presents real-world testing results of AirLoc to demonstrate its robustness and generalization ability. We collect only 4 images per room for our database and use the pretrained models described in \sref{sec:implementation} for the geometry module and NetVLAD in this demo. 
For each image in the \fref{fig:livedemo}, the left side displays the corresponding query captured by a mobile phone, while the right side shows the relocalization result.
It can be seen that AirLoc is able to relocalize well
with illumination changes in \fref{fig:illumination}, and with human interference in \fref{fig:human} as well. For better visualization, we strongly suggest the readers watch the video attached to this paper at \url{https://youtu.be/7CflVLbQOkg}.

\section{Conclusion}

In this work, we present a novel indoor relocalization method, AirLoc, which can play a crucial role in advancement of evolving applications such as augmented reality and indoor positioning using mobile phones. To be able  to quickly generalize to new environments we employ objects as the fundamental part of method. Specifically, AirLoc uses objects' appearance for relocalization and relative object geometry to differentiate between scenes having similar objects.
Our experiments show that AirLoc outperforms existing methods and achieves best perfromance on newly rendered Reloc110 dataset. We envision AirLoc to play a pivotal role in development of robust and generalizable Indoor Positioning systems for robots and humans.

\section{Acknowledgement}
This work was supported by OPPO US, the Spatial AI \& Robotics (SAIR) Lab at State University of New York at Buffalo, and the AirLab at Carnegie Mellon University.

%%%%%%%%%%%%%%%%%%%%%%%%%%%%%%%%%%%%%%%%%%%%%%%%%%%%%%%%%%%%%%%%%%%%%%%%%%%%%%%%

%%%%%%%%%%%%%%%%%%%%%%%%%%%%%%%%%%%%%%%%%%%%%%%%%%%%%%%%%%%%%%%%%%%%%%%%%%%%%%%%

% \nocite{*}  % Without this, cite articles in text using \cite{...}
\bibliographystyle{IEEEtran}
\bibliography{./IEEEfull,refs}

\end{document}